# Bi-directional Motion Detection: A Neural Intelligent Model for Perception of Cognitive Robots

Matin Macktoobian
*Fault Detection & Identification Lab (FDI)*
*Faculty of Electrical & Computer Engineering*
*K. N. Toosi University of Technology*
*Tehran, Iran*
*matinking@hotmail.com*

***Abstract*:** In this paper, a new neuronal circuit, based on the spiking neuronal network model, is proposed in order to detect the movement direction of dynamic objects wandering around cognitive robots. Capability of our new approach in bi-directional movement detection is beholden to its symmetric configuration of the proposed circuit. With due attention to magnificence of handling of blocking problems in neuronal networks such as epilepsy, mounting both excitatory and inhibitory stimuli has been taken into account. Investigations upon applied implementation of aforementioned strategy on PIONEER™ cognitive robot reveals that the strategy leads to alleviation of potential level in the sensory networks. Furthermore, investigation on intrinsic delay of the circuit reveals not only the noticeable switching rate which could be acquired but the high-efficient coupling of the circuit with the other high-speed ones.

***Index Terms - Cognitive robots, Braitenberg vehicles, Movement detector, Neuronal networks***

## I. INTRODUCTION

Investigation on mental and psychological properties of humans and animals in order to emulate different operations of the brain have been taken into account from old eras. After some seminal explorations about construction of the neuronal cells in recent decades, many researches were commenced in order to map some simple sensory concepts of the animal brains into neuronal circuits. Analysis of excitatory and inhibitory stimuli could be considered as denouements of Cajal's primary investigations on some neuronal systems of the human such as visual system [1]. Based on some breakthroughs and new accomplishments which were acquired in 1970 decade by neuroscientists and cyberneticians, especially Braitenberg great work [2], artificial neuronal network concept was recognized as the effective and practical approach to implement all ideas for simulation of neuronal mechanisms.

Different types of vehicles which were introduced by Braitenberg could be considered as an unique strategy to depict an artificial concept of different operations of the animal brain, an imitation of rudimentary mankind behaviors just based on simple stimuli acquired from circumferential environment. In following decades, various researches in different concepts related to basic Braitenberg vehicles were done. Robotics could be considered as one of the PIONEER™ sciences which tried to employ these intrinsic smart creatures in mobile robots. It is simplicity itself that control of mobile robots is a crucial stuff which needs to complete knowledge about operation of the robot. With due attention to empirical attitude which was issued upon Braitenberg vehicles in most of researches, fuzzy control was taken into account. Beaufrere's viewpoint introduced in [3] continued in next researches. A sensor-interpreting approach was implemented by Maaref et al, in [4]. Combination of the fuzzy logic with different strategies is theme of the other researches. Application of genetic algorithms and adaptive fuzzy-neural networks cooperated with fuzzy logic were studied by Pratihar in [5] and Wang et al in [6], respectively. [7] investigated by Yang et al could be considered as a successful model for navigation of differential mobile robots. A solution on navigation problem in corridor is proposed by Lee et al in [8].

French's attempts in switching between different types led to introduction of the Neuromodulation concept in Braitenberg vehicles which is investigated in [9]. Finally, the most recent accomplishments in Braitenberg vehicles territory could be referred to Rano who has tried to extract some mathematical models for stimuli in order to reach a more investigated attitude regarding some basic vehicles like vehicle 3a, which is investigated in [10].

Neuronal circuits could be implemented in huge neuronal networks in order to syncretize various sensory operations. As Braitenberg's attitude and based on the fact that different operations need different neuronal structures, in other words neuronal circuits, optimization of sensory modules could be considered as a worthwhile consequence to acquire embedded neuronal systems as much as similar to nervous construction of brain. Therefore, it sounds reasonable to segment our artificial brain into separate parts. Undoubtedly, cognition of dynamic objects around the vehicle is one of the most important senses which is desired for the creature. With taking the practical points into consideration, volume of circuit and delay of different synapses, in addition to some particular phenomena like epilepsy, must be taken into account.

Detection of movement direction, as a necessary operation in cognition process, is investigated in Braitenberg's great book and some primary designs are proposed there. Most of his sensory-designs are established based on activation of one or some neurons due to applied stimulating input.

This paper discusses a novel movement detector which in one hand is able to distinguish both movements to the right and to the left. On the other hand, simultaneous utilization of excitatory and inhibitory stimuli has a noticeable effect on decrement of possibility of epileptic seizures in the circuit.

The organization of the rest of the paper is as follows, section II includes s a brief review on some basic definitions of neuroscience which will be mentioned and investigated in the paper frequently. Section III describes the primary motion detector which is introduced by Braitenberg. Bi-directional movement detector is presented in section IV. Section V shows the comparisons and simulated results. Conclusions and probable themes for future works are investigated in section VI.

## II. SOME SEMINAL DEFINITIONS

Below terms could be evaluated as not only basic and general concepts of neuroscience but also the criteria to compare new accomplishments with prior designs.

Stimulus: A burst of action potentials to activate sensory mechanisms.

Epilepsy: In a population of elements in which excitatory connections abound, if the number of active elements reaches a certain critical level, chances are the remaining ones will also become activated. These elements, in turn, keep the first set active. A maximal condition of activity is then established and maintained until the supply of energy is exhausted.

Excitatory: Strengthening effect of the stimuli on activation potentials.

Inhibitory: Attenuating effect of the stimuli on activation potentials.

## III. BRAITENBERG'S MOVEMENT DETECTOR PROTOTYPE

Visual system in Braitenberg's for discriminating the movement detection view is constructed entirely of photocells. His primary 3-stage neuronal circuit for "left-to-right" detection is depicted as Fig.1.

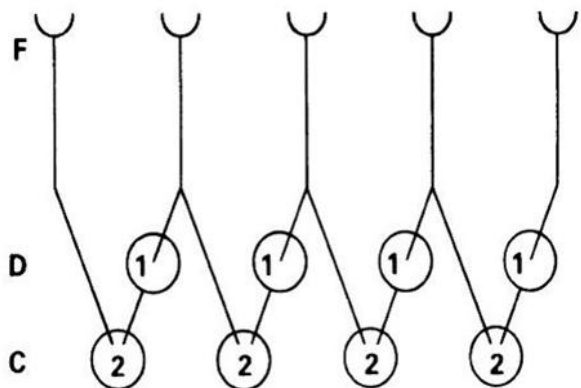

Fig 1. Braitenberg's left to right motion detector [2]

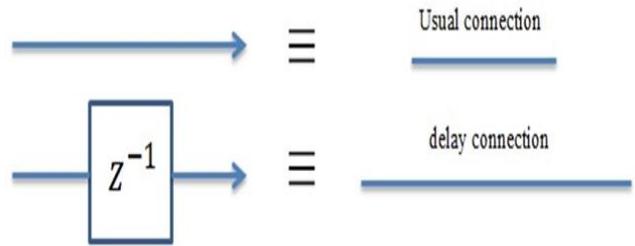

Fig 2. Typical delay model in neuronal circuits

Level (F) is responsible for receiving stimuli which are produced due to object movement. Operation of the sensors mounted in this level is solely relied on object existence in active scope of sensors. Such longer connections are in the neuronal circuit, more delay should be taken into account for those connections. Such delays are shown as simple digital filters in common notation of artificial neuronal network texts, as shown in Fig. 2. With due attention to applied delay on the left branch between each adjacent pair of sensors in level (D), exerted neuron on right branch will be activated on time so that excitatory signals from both left and right branches will activate neuron of level (C). Object movement from left to right stimulates the left-most to the right-most sensor one by one. Therefore, activation of neurons of level (C) regarding the direction is the same with the movement direction of the object. Obviously, neuronal circuit which is introduced above is not able to detect movement direction from left to right. Such circuit could be acquired with displacing of delay connection and unit excitatory neuron from left branch to the right branch and vice versa, regularly. This approach not only leads to a sophisticated module but also epilepsy could be occurred with more probability. If the vehicle is surrounded by more than one dynamic object, it is possible that both sub-circuits of the module become active simultaneously. Considering a practical view, with increasing the number of the neurons in actual systems, epileptic seizures will be evaluated as a more crucial problem.

## IV. BI-DIRECTIONAL MOVEMENT DETECTOR

Simultaneous detection of movement directions to the right or to the left requires a symmetric construction for neuronal circuit that all activation time slots must sound entirely similar. The novel movement detector is depicted as Fig. 3.

The circuit is consists of usual neurosensors and two distinctive parts for direction detection, upper one for right-to-left and the other one for left-to right. Furthermore, each part is intrinsically constructed of two stages. First stage neurons, called simple neurons, are activated by just one stimulus whereas an excitatory-inhibitory set of neurotransmmiters is embedded in order to connect the neuron to the prior and posterior neurosensors.

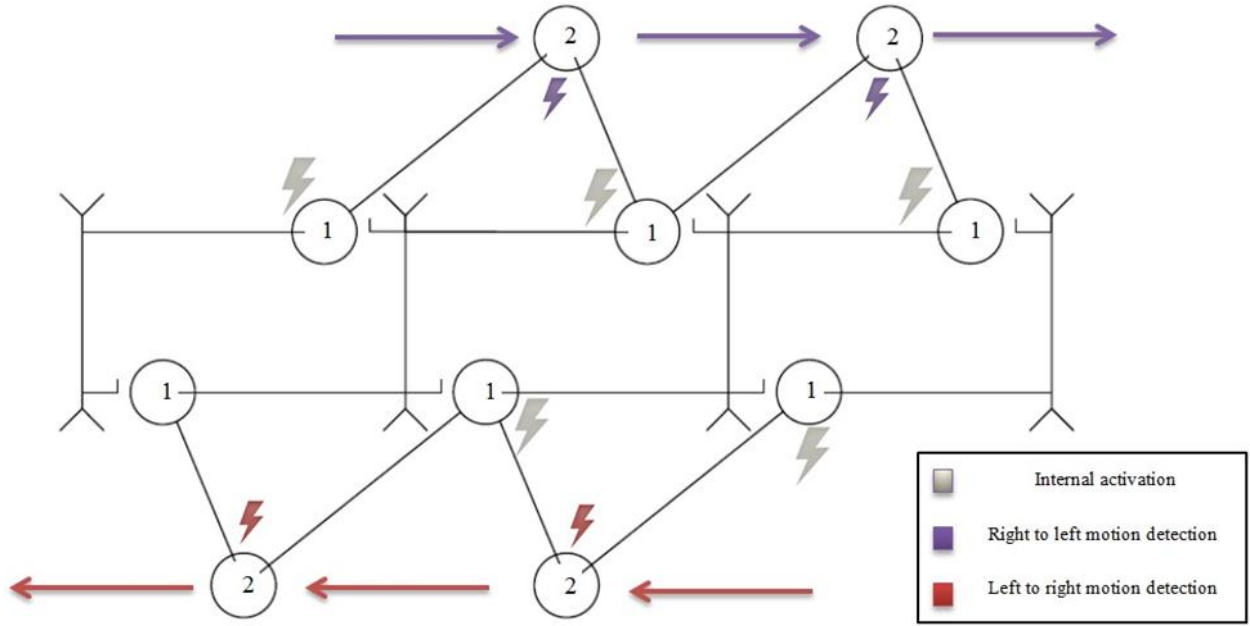

Fig 3. Bi-directional movement detection

Activation signals of each pair of successive neurons, as excitatory stimuli, are utilized to acvtivate second stage which is itself constructed of neurons, known as combined neurons, activated by two simultaneous stimuli. Unequal length of neurotransmmiters devised in both stages are necessary for overall valid operation of the circuit. With due attention to similar mechanism of both sub systems, investigation on operation of one of them could be verified the other subsystem. Here the left-to-right subcircuit is going to be explained. As a moving object approaches to a neurosensor, its activation signal will be transmitted to the simple neuron. Due to noticeable delay of excitatory connection, one can take some times into account for comprehension of activation signal by the simple neuron. When object comes rear enough to the next neurosensor and just after activation of simple neuron and transmission of its excitatory signal to ward combined sensor, inhibitory stimulus which is established by next neurosensor will deactivate the simple neuron. In addition, activation of next neurosensor also leads to activation of the anterior simple neuron with latency. Finally, shorter length of the right excitatory signal and its trivial latency will be syncretized with longer length of the left one and its noticeable latency in order to provide two desired excitatory signals for activation of corresponding combined neuron. Movement direction of the object could be evaluated by right-handed activation succession of the combined neurons. Above mechanism also could be valid for right-to-left subcircuit except the activation direction of the combined neurons which is to the left.

## V. AN INVESTIGATION ON IMPROVEMENTS

Proposed bi-directional movement detector, which is analysed in previous section, could be assessed as a better generation of the movement detectors in comparison with Braitenberg prototype model. This claim could be proved with examination of two improved factors: circuit size and epilepsy inhibition. In one hand, Given the extent of Fig. 1 and Fig.3 with consideration of using 5 neurosensors, Braitenberg's model needs 16 neurons for direction detection with mounting of two neuronal circuits individually, Whereas, bi-directional movement detector just needs 14 neurons under aegis of one circuit for providing similar operation. Fig. 4 shows
all the delay which must be taken into account for one circuit for providing similar operation. Fig. 4 shows all the delay which must be taken into account for all the delay which must be taken into account for regulation of the spikes. It is simplicity itself that delays must be controlled based in the velocity of the moving object. This quantity could be acquired by tuning of activation functions.

On the other hand, simultaneous utilization of excitatory and inhibitory connections stabilize the potential level of the system so that each set of neurons will be suppressed by reversed-type connections mounted to them.

## VI. DELAY TUNING OF CONNECTIONS

As mentioned before and based on the convention illustrated by Fig.2, longer connections are representatives of the more dilatory ones. Desired operation of the circuit depends on well-tuned latencies for combined neuron connections. It is possible to guess about important role of distance between each adjacent neurosensors and velocities of moving objects, which should be proved.

**Proposition**: If $\Delta_{RB}$ and $\Delta_{LB}$ are latency of right and left excitatory signals of combined neurons, respectively and $R$ is distance between each adjacent neurosensors

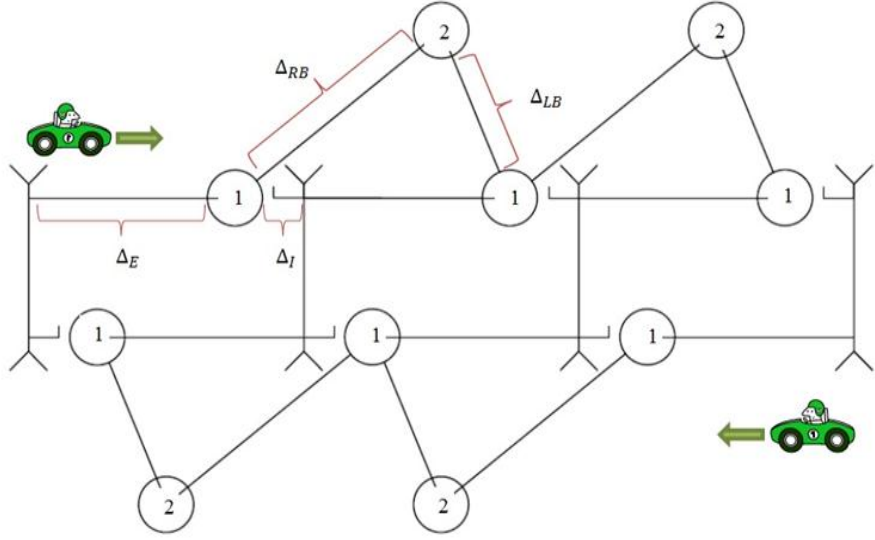

Fig. 4. Illustarion of the delays in one unit of the circuit

and $V$ is velocity of moving objects, $\Delta_{RB}$ and $\Delta_{LB}$ could be tuned by below equation:

$$\Delta_{RB} - \Delta_{LB} = \frac{R}{V}$$

**Definitions**:

$t_0$: time origin as the moving object is once seen by the most-right or most-left neurosensor.

$A_R$: activation time of right simple neuron corresponding to an unique combined neuron.

$A_L$: activation time of left simple neuron corresponding to an unique combined neuron.

$C_R$: right activation time of connection of a combined neuron.

$C_L$: left activation time of connection of a combined neuron.

**Proof**: below equations could be acquired based on Fig.4 and above definitions:

$$A_L = t_0 + \Delta E$$
$$A_R = t_0 + \Delta E + \frac{R}{V}$$
$$C_L = A_L + \Delta_{RB}$$
$$C_R = A_R + \Delta_{LB}$$

Combined neuron could be become on if only if it both activation signals receive simultaneously which means below condition must be satisfied:

$$C_R = C_L$$

So, with fulfilling above condition, the proof is completed:

$$\Delta_{RB} - \Delta_{LB} = \frac{R}{V}$$

## VII. SIMULATED DENOUEMENTS

Simulated and practical tests show noticeable accomplishments acquired by applying bi-directional movement detector on neurorobots. PIONEER™ mobile robot, shown in Fig.5, was utilized as a test bed for verifying the simulated results. It is equipped with 4 IR transducers as artificial neurosensors for this research. We evaluated the performance of bi-directional movement detection on it. Embedded IR transducers transformed it into a Braitenberg-like machine. In these experiments, the stimuli of the moving objects in front of the vehicle were modelled as Gaussian-like signals which being applied in a determined range of distances from IR transducers. With due attention to desire for testing the detector under effect of different stimuli regarding to intensity of the stimulation, moving objects were allowed to cross in front of the vehicle with various velocities and distance of the moving object from vehicle sensors, reactions of the vehicle are recorded as numerical potential quantities. The coherent denouements after raw data interpolation and computer simulations lead to reaching a rough comparison between two circuits in epilepsy rejection. Obviously, such lower the circuit potential within the experiment toward movement of dynamic objects is, more security against critical bound of epilepsy could be taken into account. Such investigation is depicted as Fig.7. With consideration the stimuli shown in Fig.6, our bi-directional movement detector works with lower potential level in comparison with Braitenberg's prototype detector. Therefore, less number of active neurons of the bi-directional movement

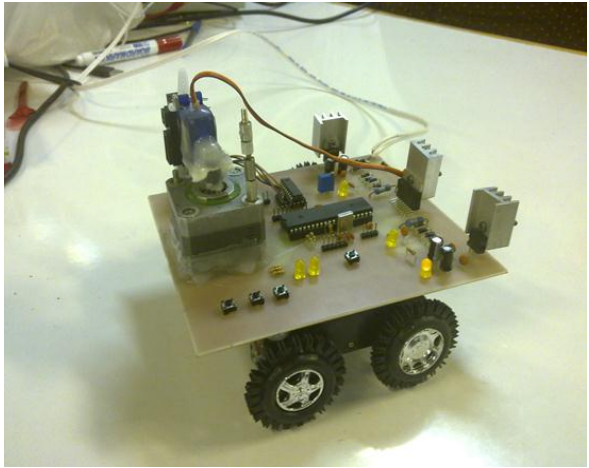

Fig 5. PIONEER™ mobile robot

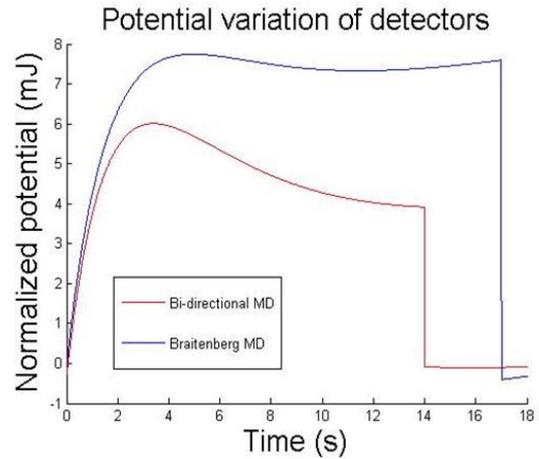

Fig 7. Potential variation of the detectors

detector in sampling time interval of the experiment could be construed, reasonably.

Spatial model of the bi-directional movement detector could be visualized as Fig. 8.

VIII. CONCLUDING REMARKS AND FUTURE WORKS

Braitenberg vehicles as one of the most inspiring concepts in various investigations on neurorobotics, have a long time to be progressed in a such complete form that different neuronal and psychological operations of the brain could be applied on them. So, robust perception of the surrounding environment could be considered as a seminal requirement for reaching to this great goal. Presented investigation in this paper shows that bi-directional movement detection not only can improve the size of the neuronal circuit, a crucial point in huge neuronal systems, but is able to ameliorate potential level of the different parts of the circuit far from danger of the epilepsy phenomenon. tuning of connections will be more complicated with consideration of multiple moving objects while they have different distances from neurosensors. Also, hardware implementation of neurorobotics systems like introduced bi-directional movement detector on some architectures like ZISC could be taken into account.

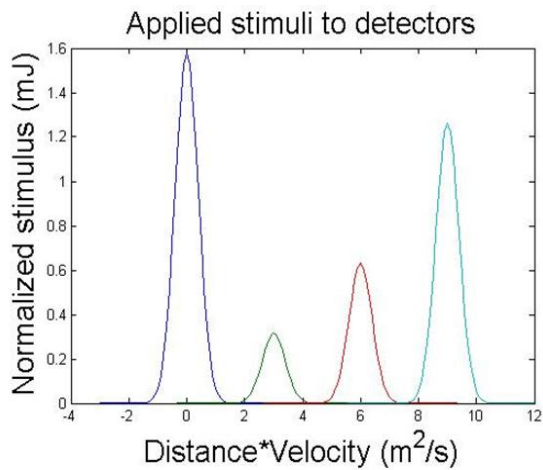

Fig 6. Applied stimuli to the detectors

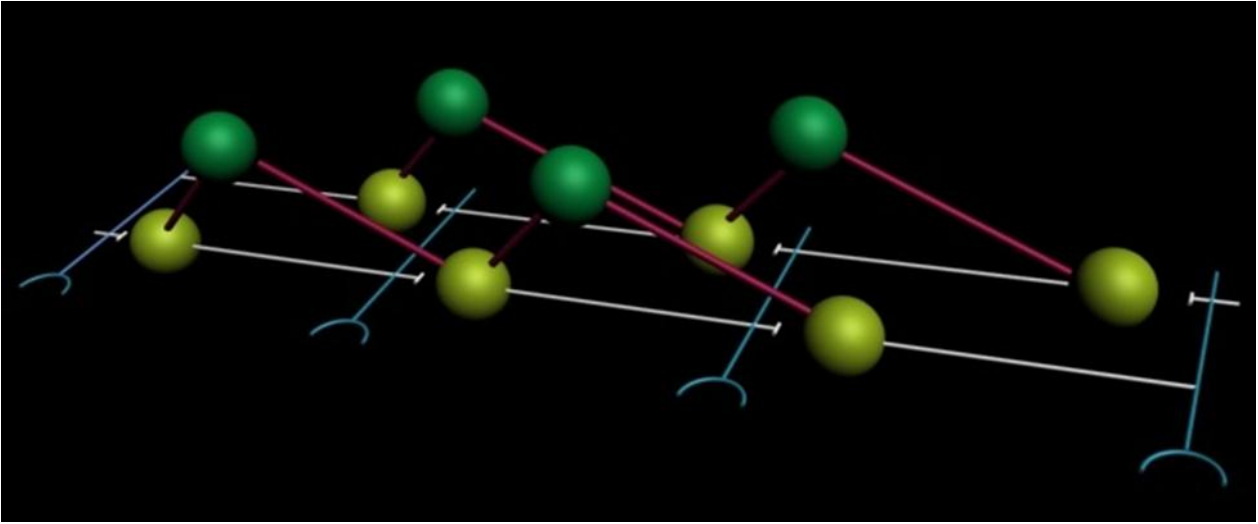

Fig. 8. 3D visualization of the bi-directional movement detector